\documentclass{article}

\usepackage[final]{nips_2016}

\usepackage[utf8]{inputenc} 
\usepackage[T1]{fontenc}    
\usepackage{hyperref}       
\usepackage{url}            
\usepackage{booktabs}       
\usepackage{amsfonts}       
\usepackage{nicefrac}       
\usepackage{microtype}      
\usepackage{xcolor}
\usepackage{latexsym}
\usepackage{graphicx}
\usepackage{amsmath}
\usepackage{bm}
\usepackage{tabularx}
\usepackage[margin=6pt,
font+=small,labelformat=parens,labelsep=space,
skip=6pt,list=false,hypcap=false
]{subcaption}

\newenvironment{q3}{\par\color{black}}{\par}

\newenvironment{q1}{\par\color{black}}{\par}

\title{C-RNN-GAN: \\ Continuous recurrent neural networks \\ with adversarial training}

\author{
  Olof Mogren\\
  Chalmers University of Technology, Sweden\\
  \texttt{olof@mogren.one} \\
}

\date{}

\begin{document}
\maketitle
\begin{abstract}
  \begin{q1}
Generative adversarial networks have been proposed as a
way of efficiently training deep generative neural networks.
We propose a generative adversarial model that works on
continuous sequential data, and apply it by training it
on a collection of classical music.
We conclude that it generates music that sounds
better and better as the model is trained, 
report statistics on generated music, and let the
reader judge the quality by downloading the generated
songs.
  \end{q1}
\end{abstract}

\section{Introduction}

Generative adversarial networks (GANs) are a class of neural
network architectures designed with the aim of generating realistic
data~\citep{goodfellow2014generative}.
The approach involves training two neural models with conflicting
objectives, one generator ($G$), and one discriminator ($D$), forcing each other to improve.
The generator tries to produce samples that looks real,
and the discriminator tries to discriminate between
generated samples and real data.
Using this framework makes it possible to train deep generative
models without expensive normalizing constants, and the technique has proven
to produce highly realistic samples of data~\citep{denton2015deep,radford2016unsupervised,im2016generating}.

In this work, we investigate the feasibility of using adversarial training
for a sequential model with continuous data, and evaluate it using 
classical music in freely available midi files.

\begin{q3}

Recurrent neural networks (RNNs) are often used to model sequences of data.
These models are usually trained using a maximum likelihood criterion.
E.g. for language modelling, they are trained to predict the next
token at any point in the sequence,
i.e. to model the conditional probability of the next token given the sequence
of preceding tokens.
By sampling from this conditional distribution, one can
generate reasonably realistic sequences~\citep{graves2013generating}.
The sampling is non-trivial and you usually resort to beam-search
to generate good sequences in tasks such as machine translation~\citep{sutskever2014sequence}.

RNNs have been used to model music~\citep{eck2002finding,boulanger2012modelling,yu2016seqgan},
but to our knowledge they always use a symbolic
representation. In contrast, our work demonstrates how one
can train a highly flexible and expressive model with fully
continuous sequence data for tone lengths, frequencies,
intensities, and timing.

We propose a recurrent neural network architecture, C-RNN-GAN
(Continuous RNN-GAN),
that is trained with adversarial training to model the whole joint probability of
a sequence, and to be able to generate sequences of data.
Our system is demonstrated by training it on
sequences of classical music in midi-format, and evaluated using
metrics such as scale consistency and tone range.

We conclude that generative adversarial training is a viable way
of training networks that model a distribution over sequences
of continuous data, and see potential for also modelling many other types of
sequential continuous data.
\end{q3}



\begin{q3}

Work using RNNs for music generation includes \citep{eck2002finding},
modelling blues songs with 25 discrete tone values, and \citep{boulanger2012modelling},
combining the RNN with restricted Boltzmann machines, representing
88 distinct tones.
\citet{yu2016seqgan} trained an RNN with adversarial training,
applying policy gradient methods to cope with the discrete
nature of the symbolic representation they employed.
%
%
In contrast to this, our work represents tones using real valued continuous
quadruplets of frequency, length, intensity, and timing.
This allows us to use standard backpropagation to train the whole model
end-to-end.
\citet{im2016generating} presented a recurrent model with adversarial training
to generate images. The LAPGAN model~\citep{denton2015deep} is another
sequential model with adversarial training, that generates images in
a coarse-to-fine fashion.
%

\end{q3}

\begin{figure}[t]
  \centering
  \includegraphics[width=.55\columnwidth]{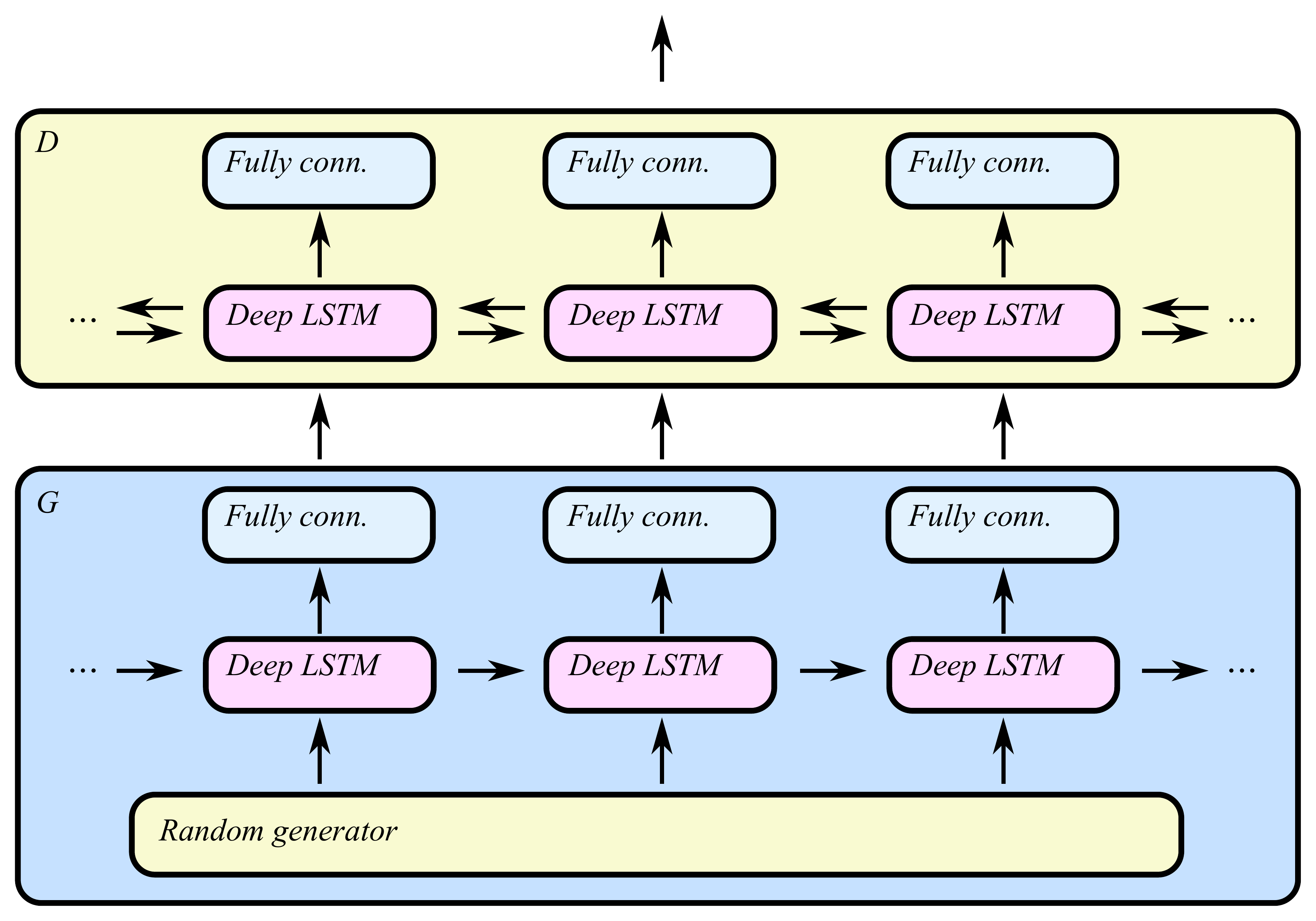}
  \caption{C-RNN-GAN. The generator ($G$) produces sequences of continuous data events. The discriminator ($D$) is trained to distinguish between real music data and generated data.}
  \label{fig:rnn-gan}
\end{figure}

\section{C-RNN-GAN: A continuous recurrent network with adversarial training}
\label{sec:c-rnn-gan}

The proposed model is a recurrent neural network with adversarial training.
The adversaries are two different deep recurrent neural models,
a generator ($G$) and a discriminator ($D$). The generator is trained
to generate data that is indistinguishable from real data, while the
discriminator is trained to identify the generated data.
The training becomes a zero-sum game for which the Nash equilibrium
is when the generator produces data that the discriminator cannot
tell from real data. We define the following loss functions $L_D$ and $L_G$:
\begin{align*}
L_G &= \frac{1}{m} \sum_{i=1}^m \log(1-D(G(\boldsymbol{z}^{(i)}))) \\
L_D &= \frac{1}{m} \sum_{i=1}^m \left[ -\log D(\boldsymbol{x}^{(i)}) - (\log(1-D(G(\boldsymbol{z}^{(i)})))) \right]
\end{align*}
(where $\boldsymbol{z}^{(i)}$ is a sequence of uniform random vectors in $[0,1]^k$, and $\boldsymbol{x}^{(i)}$ is a sequence from the training data. $k$ is the dimensionality of the data in the random sequence.)


The input to each cell in $G$ is a random vector,
concatenated with the output of previous cell.
Feeding the output from the previous cell
is common practice when training RNNs as language models~\citep{mikolov2010recurrent},
and has also been used in music composition~\citep{eck2002finding}.


The discriminator consists of a bidirectional recurrent net,
allowing it to take context in both directions into account
for its decisions.
In this work, the recurrent network used is the
Long short-term memory (LSTM)~\citep{schmidhuber1997long}.
It has an internal structure with gates that help with the vanishing
gradient problem, and to learn longer dependencies~\citep{hochreiter1998vanishing,bengio1994learning}.

\subsection{Modelling music}

In this work, 
we set out to evaluate
the viability of using generative adversarial models to learn the generating
distribution behind classical music.
Inspired by the venerable MIDI format for communicating signals between
digital musical instruments, we model the signal with four real-valued
scalars at every data point: \textit{tone length}, \textit{frequency},
\textit{intensity}, and \textit{time} spent since the previous tone.
Modelling the data in this way allows the network to represent
polyphonous chords (with zero time between two tones).
To evaluate its effect on polyphony, we have also experimented with
having up to three tones represented
as output from each LSTM cell in $G$
(with corresponding modifications to $D$).
Each tone is then represented with its own quadruplet of values as described above.
We refer to this version as \textbf{C-RNN-GAN-3}.
Similarly to the MIDI format, the absence of a tone is
represented by zero intensity output.



\section{Experimental setup}




\textbf{Model layout details:} The LSTM network in both $G$ and $D$ has depth $2$, each LSTM cell has 350 internal (hidden) units.
$D$ has a bidirectional layout, while $G$ is unidirectional.
The output from each LSTM cell in $D$ are fed into a fully connected layer
with weights shared across time steps, and one sigmoid output per cell is then averaged
to the final decision for the sequence.

\textbf{Baseline:} Our baseline is a recurrent network similar to our generator,
but trained entirely to predict the next tone event at each
point in the recurrence.


\textbf{Dataset:} Training data was collected from the web in the form of
music files in midi format, containing well-known
works of classical music.
Each midi event of the type ``note on'' was loaded and saved together with
its duration, tone, intensity (velocity), and time since beginning of last tone.
The tone data is internally represented with the corresponding sound frequency.
Internally, all data is normalized to a tick resolution of 384 per quarter note.
The data contains 3697 midi files from 160 different composers of classical music.

The source code is available on Github\footnote{\url{https://github.com/olofmogren/c-rnn-gan}},
including a utility to download
all data used in this study from different websites.



\textbf{Training:} Backpropagation through time (BPTT) and mini-batch stochastic gradient descent was used. Learning rate was set to 0.1, and we apply L2 regularization to the weights both in $G$ and $D$.
The model was pretrained for 6 epochs with a squared error loss for
predicting the next event in the training sequence.
Just as in the adversarial setting, the input to each
LSTM cell is a random vector $\boldsymbol{v}$, concatenated with the
output at previous time step. $\boldsymbol{v}$ is uniformly distributed in
$[0,1]^k$, and $k$ was chosen to be the number of features
in each tone, 4.
During pretraining, we used a schema for sequence length, beginning with short sequences, randomly sampled from the training data, eventually training the model with increasingly long sequences. We consider this a form of curriculum learning, where we begin by learning short passages, and the relations between points that are near in time.

During adversarial training, we noticed that $D$ can become too strong, resulting in a gradient that cannot be used to improve $G$.
This effect is particularly clear when the network is initialized without pretraining.
For this reason, we apply \textit{freezing}~\citep{salimans2016improved}, which means stopping the updates of $D$ whenever its training loss is less than 70\% of the training loss of $G$. We do the corresponding thing when $G$ become too strong.


We also employ \textit{feature matching}~\citep{salimans2016improved},
an approach to encourage greater variance in $G$, and
avoid overfitting to the current discriminator
by replacing the standard generator loss, $L_G$.
Normally, the objective for the generator is to maximize
the error that the discriminator makes, but with feature matching,
the objective is instead to produce an internal representation
at some level in the discriminator that
matches that of real data.
We choose the representations $R$ from the last layer before the
final logistic classification layer in $D$, and define the new
objective $\hat{L_G}$ for $G$.
\[ \hat{L_G} = \frac{1}{m} \sum_{i=1}^m (R(\boldsymbol{x}^{(i)})-R(G(\boldsymbol{z}^{(i)})))^2  \]


\begin{figure}[ht!]
  \centering
  \includegraphics[width=.80\columnwidth]{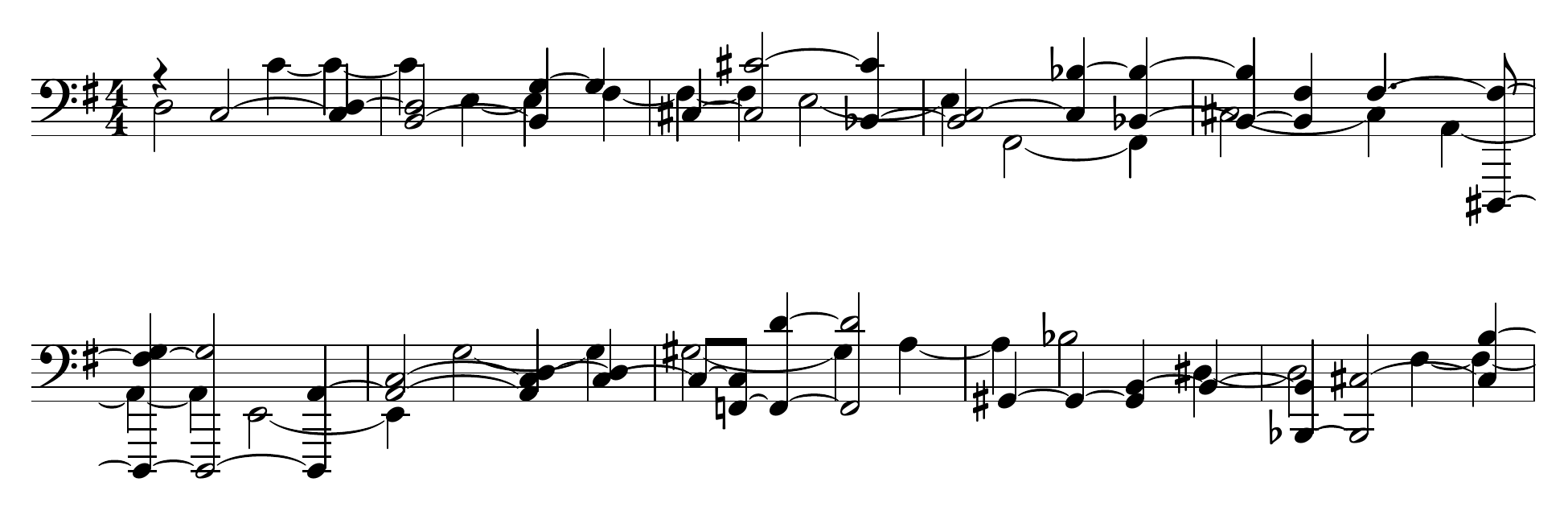}
  \caption{Music generated with C-RNN-GAN with feature matching and three tones per cell.}
  \label{fig:c-rnn-gan}
\end{figure}

\begin{figure}[ht!]
  \centering
\begin{subfigure}[ht]{.5\textwidth}
  \centering
  \includegraphics[width=.95\linewidth]{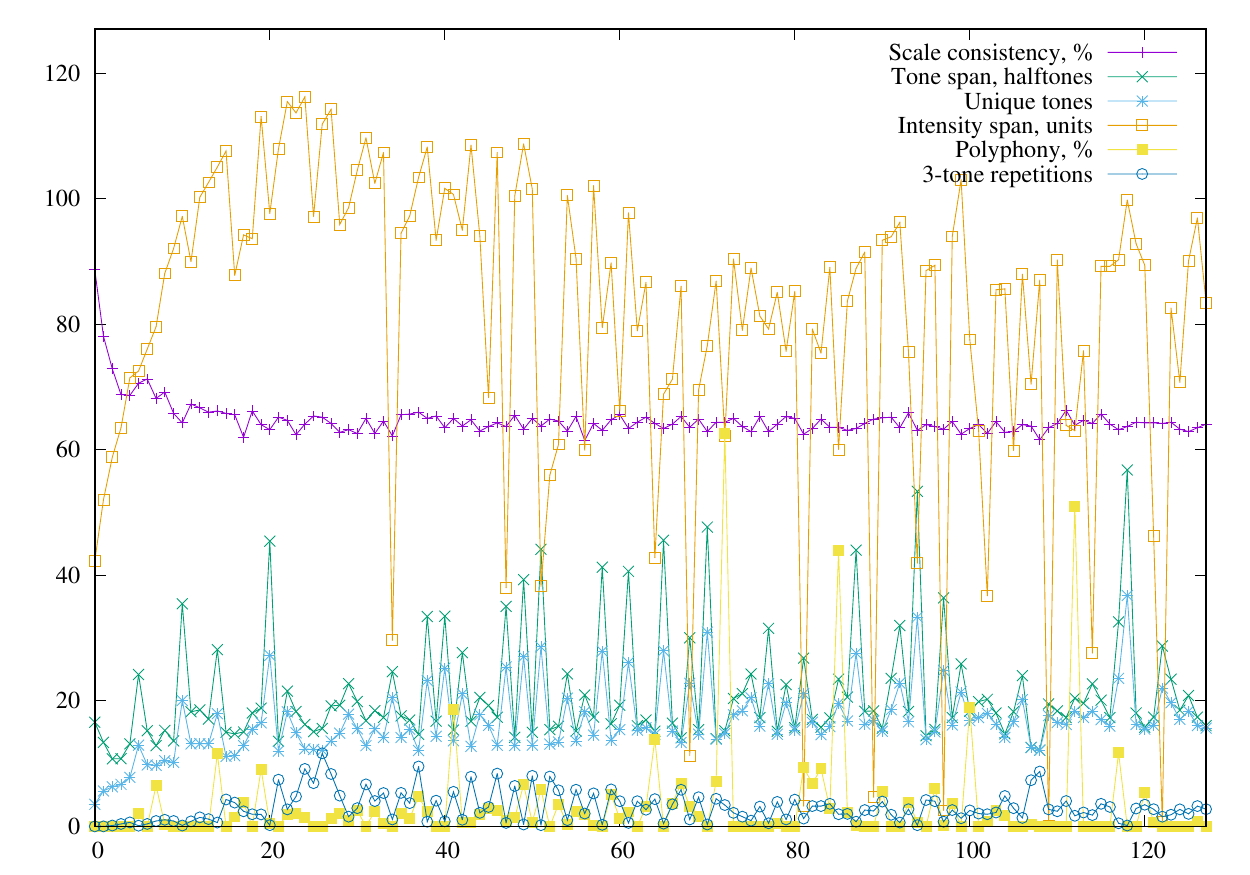}
\caption{
\textbf{C-RNN-GAN} using feature matching.
}
  \label{fig:results-c-rnn-gan}
\end{subfigure}%
\begin{subfigure}[ht]{.5\textwidth}
  \centering
  \includegraphics[width=.95\linewidth]{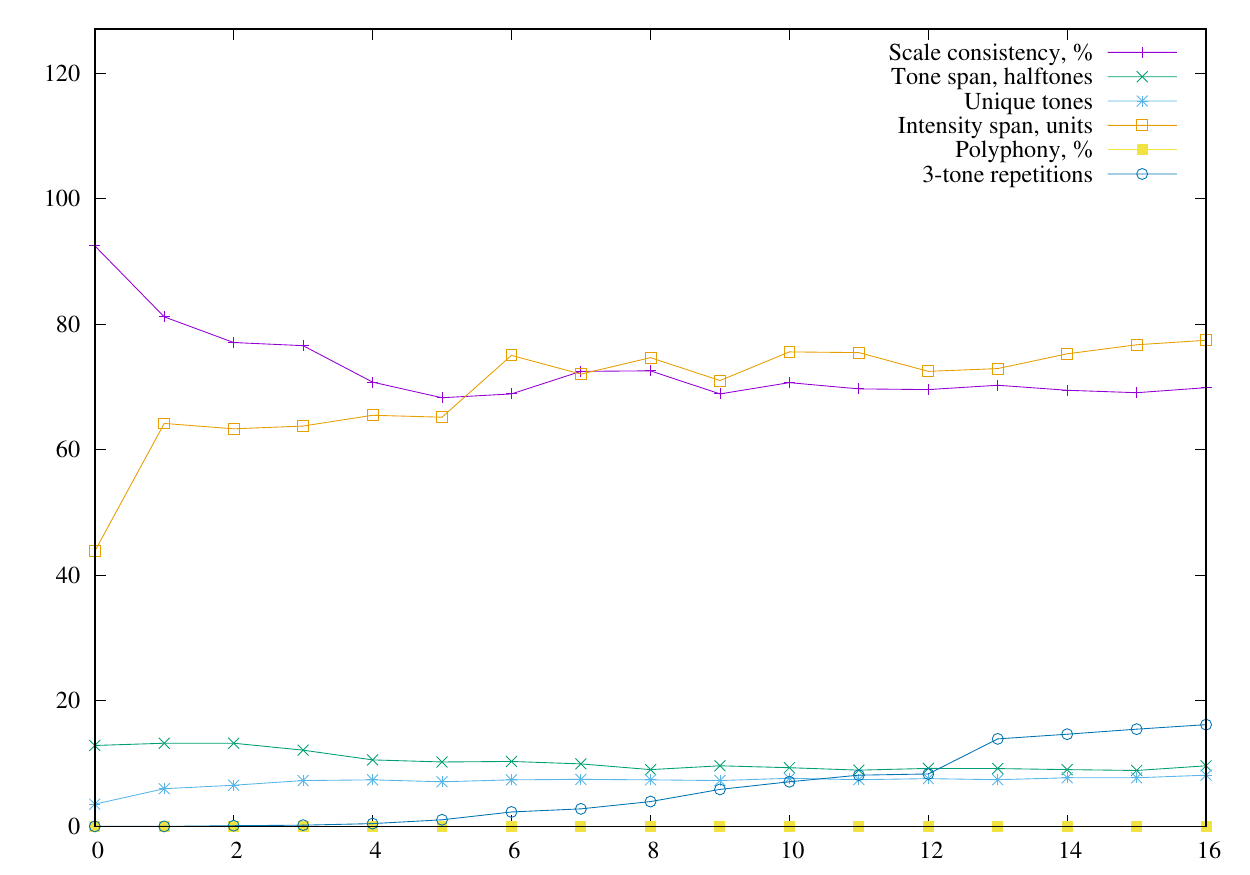}
\caption{\textbf{Baseline} with maximum likelihood training.
}
  \label{fig:results-baseline}
\end{subfigure}
\begin{subfigure}[ht]{.5\textwidth}
  \centering
  \includegraphics[width=.95\linewidth]{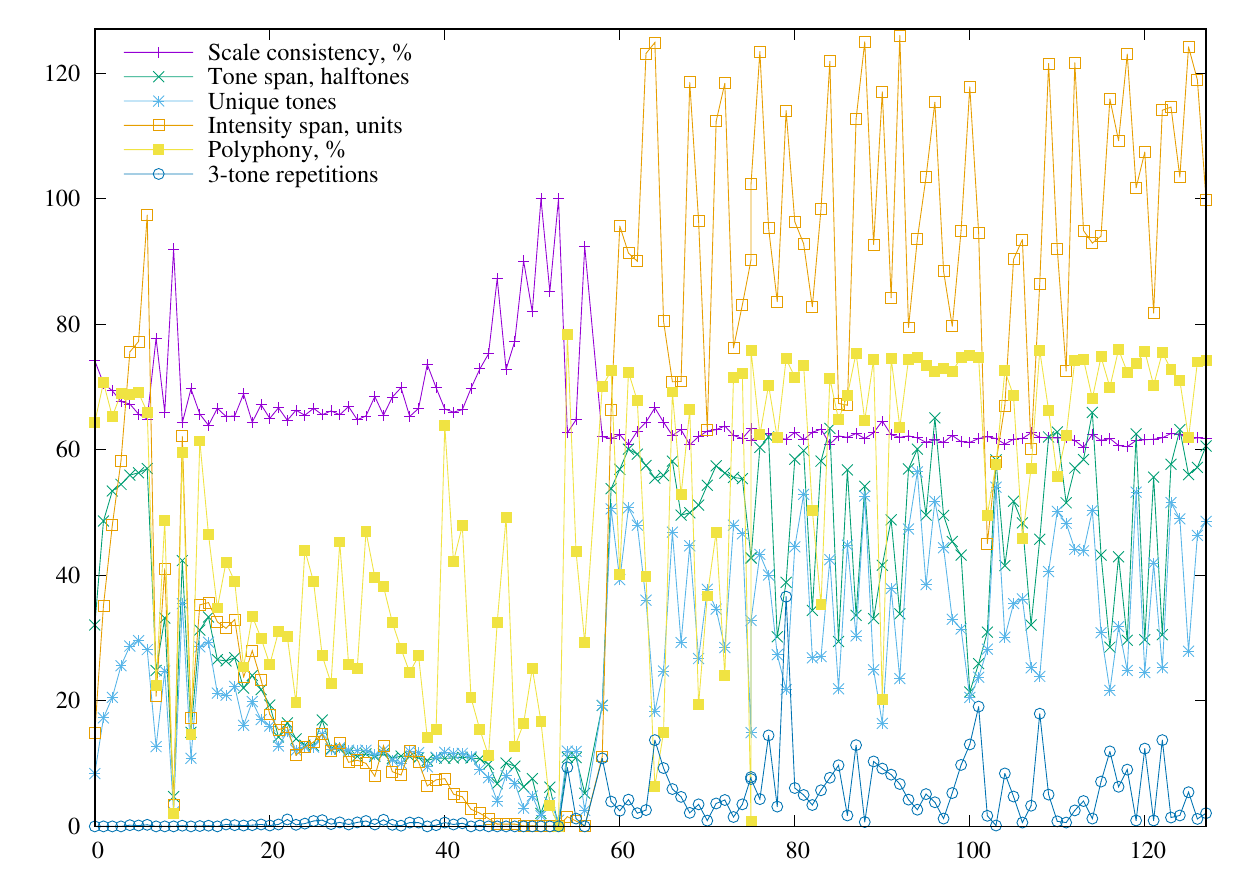}
\caption{
\textbf{C-RNN-GAN-3}, using feature matching
and three tone outputs per LSTM cell.
}
\label{fig:results-c-rnn-gan-3}
\end{subfigure}%
\begin{subfigure}[ht]{.5\textwidth}
  \centering
  \includegraphics[width=.95\linewidth]{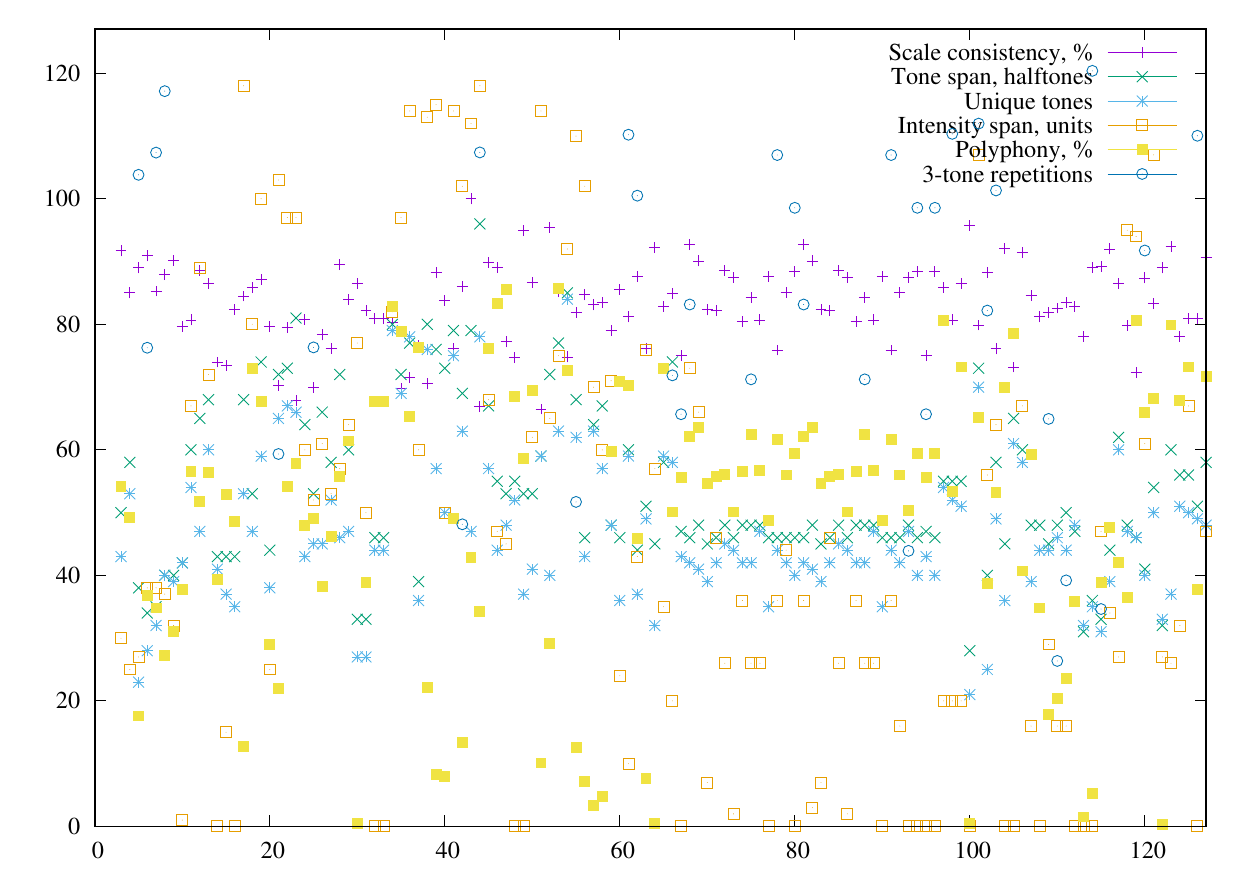}
\caption{
The same statistics for a selection of real music in the dataset.
}
\label{fig:results-real-music}
\end{subfigure}
\caption{
Statistics of generated music from the evaluated models.
C-RNN-GAN (\subref{fig:results-c-rnn-gan}) generates
music with increasing complexity as training proceeds.
The number of unique tones used is has a vaguely increasing trend,
while the scale consistency seems to stabilize after ten or fifteen epochs.
The 3-tone repetition has an increasing trend the first 25 epochs,
and then stays on quite a low level, seemingly correlated with
the number of tones used.
The baseline model (\subref{fig:results-baseline})
does not reach the same level of variation.
The number of unique tones used is consistently much lower
while the scale consistency seems to be similar to C-RNN-GAN.
The tone span follows number of unique tones more closely than with
C-RNN-GAN, suggesting that the baseline has less variability in the tones used.
The C-RNN-GAN-3 (\subref{fig:results-c-rnn-gan-3})
obtains a higher polyphony score,
in contrast to both C-RNN-GAN and the baseline.
After reaching a state with many zero-valued outputs around epoch 50 to 55,
C-RNN-GAN-3 reaches substantially higher values on tone span, number
of unique tones, intensity span, and 3 tone repetitions.
In (\subref{fig:results-real-music}), one can see that real music
has an intensity span similar
to that of the generated music. Scale consistency is slightly higher, but also
varies more. The polyphony score is similar to C-RNN-GAN-3. 3-tone
repetitions are much higher, but is difficult to compare as the songs are
of different length. The count is normalized by dividing by $l_r/l_g$,
where $l_r$ is the length of the real music, and $l_g$ is the length of the
generated music.
}
\label{fig:results}
\end{figure}

\subsection{Evaluation}
\label{sec:evaluation}

Evaluation of C-RNN-GAN was done using a number of measurements on 
generated output.

\begin{q3}


\textbf{Polyphony}, measuring how often (at least) two tones are played
simultaneously (their start time is exactly the same).
Note that this is a rather restricted metric, as it can give a low score
to music that has simultaneous tones that does not start at exactly the same time.

    
\textbf{Scale consistency} were computed by counting the fraction of
tones that were part of a standard scale, and reporting the number
for the best matching such scale.

\textbf{Repetitions} of short subsequences were counted, giving a score
on how much recurrence there is in a sample. This metric takes only the
tones and their order into account, not their timing.

\textbf{Tone span} is the number of half-tone steps between the
lowest and the highest tone in a sample.

\end{q3}

A tool was implemented to compute these estimates, which is available
on Github\footnote{\url{https://github.com/olofmogren/c-rnn-gan}}
together with all source code
used in this work.

\section{Results}
The results of the experimental study is presented in Figure~\ref{fig:results}. 
Adversarial training helps the model learn patterns with more variability,
larger tone span, and larger intensity span. Allowing the model to output
more than one tone per LSTM cell helps to generate music with a higher polyphony score.

\subsection{Listening impressions}

While we have not yet performed a thorough listening study of this work,
the impressions of the author and co-workers is that
the model trained with feature matching gets a better trade-off
between structure and surprise than the other variants.
The versions with only maximum-likelihood pretraining tends
to not give enough surprise, and the versions with mini-batch features
tended to sample music with too little structure to be interesting
to a real listener. This is a bit surprising and something that
we intend to look more into.

Music files generated with C-RNN-GAN can be downloaded from 
\mbox{\url{http://mogren.one/publications/2016/c-rnn-gan/}}.

\section{Discussion and conclusions}

In this paper, we have proposed a recurrent neural model for
continuous data, trained using an approach based on
generative adversarial networks.
While more experimentation needs to be done, we believe that
the results are promising.
We have noted that adversarial training helps the recurrent
neural network generate music that varies more in both
the number of tones used, and the span of intensities of
played tones.
The generated music can not yet compare to the music
in the training data, by human judgement.
The reasons for this remain to be explored.
However, in the evaluation (see Figure~\ref{fig:results}) one can
see that the scores of music generated using C-RNN-GAN
show more resemblance to scores of real music, than do those of
music generated using the baseline.
The generated music is polyphonous, but in the polyphony score in our experimental
evaluation, measuring how often two tones are played at exactly the same time,
C-RNN-GAN scored low. Allowing each LSTM cell to output up to three
tones at the same time resulted in a model that scored much better
with polyphony.
%
One can hear in the generated samples, that while timing
can vary quite a bit from sample to sample, it is generally
rather consistent within one track, giving a feeling of tempo in
the generated songs.



\section*{Acknowledgments}


This work has been done within the project ``Data-driven secure business intelligence'', grant IIS11-0089 from the Swedish Foundation for Strategic Research (SSF).

\clearpage




\bibliography{mogren2016c-rnn-gan}

\begin{thebibliography}{14}
\providecommand{\natexlab}[1]{#1}
\providecommand{\url}[1]{\texttt{#1}}
\expandafter\ifx\csname urlstyle\endcsname\relax
  \providecommand{\doi}[1]{doi: #1}\else
  \providecommand{\doi}{doi: \begingroup \urlstyle{rm}\Url}\fi

\bibitem[Alec~Radford(2016)]{radford2016unsupervised}
Soumith~Chintala Alec~Radford, Luke~Metz.
\newblock Unsupervised representation learning with deep convolutional
  generative adversarial networks.
\newblock In \emph{International Conference on Learning Representations}, 2016.

\bibitem[Bengio et~al.(1994)Bengio, Simard, and Frasconi]{bengio1994learning}
Yoshua Bengio, Patrice Simard, and Paolo Frasconi.
\newblock Learning long-term dependencies with gradient descent is difficult.
\newblock \emph{Neural Networks, IEEE Transactions on}, 5\penalty0
  (2):\penalty0 157--166, 1994.

\bibitem[Denton et~al.(2015)Denton, Chintala, Fergus, et~al.]{denton2015deep}
Emily~L Denton, Soumith Chintala, Rob Fergus, et~al.
\newblock Deep generative image models using a laplacian pyramid of adversarial
  networks.
\newblock In \emph{Advances in neural information processing systems}, pages
  1486--1494, 2015.

\bibitem[Eck and Schmidhuber(2002)]{eck2002finding}
Douglas Eck and Juergen Schmidhuber.
\newblock Finding temporal structure in music: Blues improvisation with lstm
  recurrent networks.
\newblock In \emph{Neural Networks for Signal Processing, 2002. Proceedings of
  the 2002 12th IEEE Workshop on}, pages 747--756. IEEE, 2002.

\bibitem[Goodfellow et~al.(2014)Goodfellow, Pouget-Abadie, Mirza, Xu,
  Warde-Farley, Ozair, Courville, and Bengio]{goodfellow2014generative}
Ian Goodfellow, Jean Pouget-Abadie, Mehdi Mirza, Bing Xu, David Warde-Farley,
  Sherjil Ozair, Aaron Courville, and Yoshua Bengio.
\newblock Generative adversarial nets.
\newblock In \emph{Advances in Neural Information Processing Systems}, pages
  2672--2680, 2014.

\bibitem[Graves(2013)]{graves2013generating}
Alex Graves.
\newblock Generating sequences with recurrent neural networks.
\newblock \emph{arXiv preprint arXiv:1308.0850}, 2013.

\bibitem[Hochreiter(1998)]{hochreiter1998vanishing}
Sepp Hochreiter.
\newblock The vanishing gradient problem during learning recurrent neural nets
  and problem solutions.
\newblock \emph{International Journal of Uncertainty, Fuzziness and
  Knowledge-Based Systems}, 6\penalty0 (02):\penalty0 107--116, 1998.

\bibitem[Im et~al.(2016)Im, Kim, Jiang, and Memisevic]{im2016generating}
Daniel~Jiwoong Im, Chris~Dongjoo Kim, Hui Jiang, and Roland Memisevic.
\newblock Generating images with recurrent adversarial networks.
\newblock \emph{arXiv preprint arXiv:1602.05110}, 2016.

\bibitem[Mikolov et~al.(2010)Mikolov, Karafi{\'a}t, Burget, Cernock{\`y}, and
  Khudanpur]{mikolov2010recurrent}
Tomas Mikolov, Martin Karafi{\'a}t, Lukas Burget, Jan Cernock{\`y}, and Sanjeev
  Khudanpur.
\newblock Recurrent neural network based language model.
\newblock In \emph{Interspeech}, volume~2, page~3, 2010.

\bibitem[Nicolas Boulanger-Lewandowski(2012)]{boulanger2012modelling}
Pascal~Vincent Nicolas Boulanger-Lewandowski, Yoshua~Bengio.
\newblock Modeling temporal dependencies in high-dimensional sequences:
  Application to polyphonic music generation and transcription.
\newblock In \emph{Proceedings of the 29th International Conference on Machine
  Learning (ICML)}, page 1159–1166, 2012.

\bibitem[Salimans et~al.(2016)Salimans, Goodfellow, Zaremba, Cheung, Radford,
  and Chen]{salimans2016improved}
Tim Salimans, Ian Goodfellow, Wojciech Zaremba, Vicki Cheung, Alec Radford, and
  Xi~Chen.
\newblock Improved techniques for training gans.
\newblock In \emph{Advances in Neural Information Processing Systems}, pages
  2226--2234, 2016.

\bibitem[Schmidhuber and Hochreiter(1997)]{schmidhuber1997long}
J{\"u}rgen Schmidhuber and Sepp Hochreiter.
\newblock Long short-term memory.
\newblock \emph{Neural computation}, 7\penalty0 (8):\penalty0 1735--1780, 1997.

\bibitem[Sutskever et~al.(2014)Sutskever, Vinyals, and
  Le]{sutskever2014sequence}
Ilya Sutskever, Oriol Vinyals, and Quoc~V Le.
\newblock Sequence to sequence learning with neural networks.
\newblock In \emph{Advances in neural information processing systems}, pages
  3104--3112, 2014.

\bibitem[Yu et~al.(2016)Yu, Zhang, Wang, and Yu]{yu2016seqgan}
Lantao Yu, Weinan Zhang, Jun Wang, and Yong Yu.
\newblock Seqgan: Sequence generative adversarial nets with policy gradient.
\newblock \emph{arXiv preprint arXiv:1609.05473}, 2016.

\end{thebibliography}
\bibliographystyle{plainnat}

\clearpage

\end{document}